\documentclass[10pt,twocolumn,letterpaper]{article}

\usepackage{cvpr}
\usepackage{times}
\usepackage{epsfig}
\usepackage{graphicx}
\usepackage{amsmath}
\usepackage{amssymb}
\usepackage{subcaption}
\usepackage{multirow}

\usepackage[pagebackref=true,breaklinks=true,letterpaper=true,colorlinks,bookmarks=false]{hyperref}

\cvprfinalcopy 

\def\cvprPaperID{****} 

\ifcvprfinal\fi
\begin{document}

\title{Attribute-Centered Loss for Soft-Biometrics Guided Face Sketch-Photo Recognition}

\author{Hadi Kazemi \hspace{0.5cm} Sobhan Soleymani \hspace{0.5cm} Ali Dabouei \hspace{0.5cm} Mehdi Iranmanesh \hspace{0.5cm} Nasser M. Nasrabadi\\
	West Virginia University\\
	{\tt\small \{hakazemi, ad0046, ssoleyma, seiranmanesh\}@mix.wvu.edu, nasser.nasrabadi@mail.wvu.edu}
}

\maketitle
\thispagestyle{empty}
\begin{abstract}
   Face sketches are able to capture the spatial topology of a face while lacking some facial attributes such as race, skin, or hair color. Existing sketch-photo recognition approaches have mostly ignored the importance of facial attributes. In this paper, we propose a new loss function, called attribute-centered loss, to train a Deep Coupled Convolutional Neural Network (DCCNN) for facial attribute guided sketch to photo matching. Specifically, an attribute-centered loss is proposed which learns several distinct centers, in a shared embedding space, for photos and sketches with different combinations of attributes. The DCCNN simultaneously is trained to map photos and pairs of testified attributes and corresponding forensic sketches around their associated centers, while preserving the spatial topology information. Importantly, the centers learn to keep a relative distance from each other, related to their number of contradictory attributes. Extensive experiments are performed on composite (E-PRIP) and semi-forensic (IIIT-D Semi-forensic) databases. The proposed method significantly outperforms the state-of-the-art.
\end{abstract}

\section{Introduction}
Automatic face sketch-to-photo identification has always been an important topic in computer vision and machine learning due to its vital applications in law enforcement \cite{wang2009recognition, liu2005nonlinear}. In criminal and intelligence investigations, in many cases, the facial photograph of a suspect is not available, and a forensic hand-drawn or computer generated composite sketch following the description provided by the testimony of an eyewitness is the only clue to identify possible suspects. Therefore, an automatic matching algorithm is needed to quickly and accurately search the law enforcements face databases or surveillance cameras using a forensic sketch. However, the forensic or composite sketches contain only some basic information of the suspects' appearance such as the spatial topology of their faces while other soft biometric traits, e.g. skin, race, or hair color, are left out. 

Traditional sketch recognition algorithms can be classified into two categories, namely \textit{generative} and \textit{discriminative} approaches. Generative approaches transfer one of the modalities into the other before matching \cite{ouyang2016forgetmenot, wang2009face, gao2008face}. On the other hand, discriminative approaches perform feature extraction, such as the scale-invariant feature transform (SIFT) \cite{klare2010sketch}, Weber's local descriptor (WLD) \cite{bhatt2012memetically}, and multi-scale local binary pattern (MLBP) \cite{galea2016face}. However, these features are not always quite optimal for a cross-modal recognition task \cite{zhang2011coupled}. As a consequence, some other methods are investigated in order to learn or extract modality-invariant features \cite{klare2013heterogeneous, huang2013coupled}. More recently, deep learning based approaches have emerged as potentially viable techniques to tackle the cross-domain face recognition problem by learning a common latent embedding between the two modalities \cite{galea2017forensic, nagpal2017face, iranmanesh18}. However, utilizing deep learning techniques for sketch-to-photo recognition is more challenging than other single modality domains since they require a large number of data samples to avoid over-fitting and local minima. Furthermore, current publicly available sketch-photo datasets comprise only a few number of sketch-photo pairs. More importantly, there is one sketch per subject in most datasets and this makes it a difficult, and sometimes impossible task for the network to learn robust latent features \cite{galea2017forensic}. As a result, most approaches have utilized a relatively shallow network, or trained the network only on one of the modalities (typically the face photo) \cite{mittal2015composite}.

Existing state-of-the-art approaches focus primarily on closing the semantic representation of the two domains whilst ignoring the absence of soft-biometric information in the sketch modality. Given the impressive results of recent sketch-photo recognition algorithms, there is still a missing part in this process which is conditioning the matching process on the soft biometric traits. Especially in the application of sketch-photo recognition, based on the quality of sketches, there are usually some facial attributes which are missing in the sketch domain, such as skin, hair, and eye colors, gender, and ethnicity. Furthermore, conditioning the matching process to other adhered facial characteristics, such as having eyeglasses or a hat, provides extra information about the individual of interest and can result in more accurate and impressive outcomes. Describing and manipulating attributes from face images have been active research topics for years \cite{zhong2016face, jourabloo2015attribute, tokola20153d}. The application of soft biometric traits in person identification has also been studied in the literature \cite{dantcheva2016else, hadi18}. 

Despite the evidence for the usefulness of facial attributes, development of a paradigm to exploit them in the sketch-photo matching has not been adequately studied. A direct suspect identification scheme based solely on descriptive facial attributes is proposed in \cite{klare2014suspect} that completely bypassed the sketch images. Klare \etal \cite{klare2011matching} utilized race and gender to narrow down the gallery of mugshots for more accurate matching.  In a more recent work \cite{ouyang2014cross}, a CCA subspace is learned to fuse the attributes and low-level features. However, they extract the features which are common in both modalities. Mittal \etal \cite{mittal2017composite} utilized the facial attributes such as gender, ethnicity, and skin color to reorder the ranked list. They also fused multiple sketches of a suspect to increase the performance of their algorithm. 

In this work, we propose a facial attribute-guided sketch-photo recognition scheme conditioned on relevant facial attributes. We introduce a new loss function, called attribute-centered loss, to capture the similarity of identities that have the same combination of facial attributes. The key element of this loss function is assigning a distinct centroid (center point), in the embedding space, to different combinations of facial attributes. To train a deep neural network using the attribute-centered loss, a pair of sketch-attribute need to be provided to the network instead of a single sketch. Our proposed loss function then encourages a deep coupled neural network to map a photo and its corresponding sketch-attribute pair as close as possible to each other in the shared latent sub-space. Simultaneously, the distance of all the photos and sketch-attribute pairs to their corresponding centers must not be more than a pre-specified margin. This helps the network to learn and filter out the subjects which are very similar in facial structure to the suspect but do not share a considerable number of attributes. Finally, the centers are trained to keep a distance related to their number of contradictory attributes. The justification behind the latter is that it is more likely that a victim misclassifies a few facial attributes of the suspect than most of them. In summary, the main contributions of this paper include the following:
\begin{itemize}
  \item We propose a novel framework for facial attribute guided sketch-photo recognition.
  \item We introduce a new loss function, namely attribute-centered loss, to fuse the facial attributes provided by eyewitnesses and the geometrical properties of forensic sketches to improve the performance of our sketch-photo recognition.
  \item The proposed loss function uses the provided attributes in a soft manner. In other words, suspects with a few contradictory attributes compared to the facial attributes described by an eyewitness can still be detected as the person of interest if their geometrical properties still have a high matching score with the forensic sketch. 
\end{itemize}

\section{Methodology}
In this section, we describe our approach. We first introduce the center loss in its general form proposed by Wen \etal \cite{wen2016discriminative}. Inspired by their work, we propose the attribute-centered loss to exploit facial attributes in sketch-photo recognition, followed by the training methodology to learn a common latent feature space between the two modalities. 
\subsection{Center Loss}
The common approach to train a deep neural network for classification or verification task is using cross-entropy loss. However, this loss function does not encourage the network to extract discriminative features and only guarantees their separability \cite{wen2016discriminative}. The intuition behind the center loss is that the cross-entropy loss does not force the network to learn the intra-class variations in a compact form. To overcome this issue, contrastive loss \cite{hadsell2006dimensionality} and triplet loss \cite{schroff2015facenet} have emerged in the literature to capture a more compact form of the intra-class variations. Despite their recent diverse successes, their convergence rates are quite slow.
Consequently, a new loss function, namely center loss, has been proposed in \cite{wen2016discriminative} to push the neural network to distill a set of features with more discriminative power. The center loss, $L_c$, is formulated as \begin{align}
\label{center_loss}
L_c = \dfrac{1}{2}\sum_{i=1}^{m}\parallel x_i - c_{y_i} \parallel_2^2, 
\end{align}
where $m$ denotes the number of samples in a mini-batch, $x_i \in {\rm I\!R}^d$ denotes the $i$th sample feature embedding, belonging to the class $y_i$. The $c_{y_i} \in {\rm I\!R}^d$ denotes the $y_i$th class center of the embedded features, and $d$ is the feature dimension. To train a deep neural network, a joint supervision of the proposed center loss and cross-entropy loss is adopted:
\begin{align}
\label{total_center_loss}
L = L_s + \lambda L_c, 
\end{align}
where $L_s$ is the softmax loss (cross-entropy). The center loss, as defined in Eq. \ref{center_loss}, is deficient in that it only penalizes the compactness of intra-class variations without considering the inter-class separation. Therefore, to address this issue, a contrastive-center loss has been proposed in \cite{qi2017contrastive} as
\begin{align}
\label{contrasive_center_loss}
L_{ct-c} = \dfrac{1}{2}\sum_{i=1}^{m}\dfrac{\parallel x_i - c_{y_i} \parallel_2^2}{(\sum_{j=1, j\neq y_i}^{k} \parallel x_i - c_j \parallel_2^2)+\delta},
\end{align}
where $\delta$ is a constant preventing a zero denominator, and $k$ is the number of classes. This loss function not only penalizes the intra-class variations but also maximizes the distance between each sample and all the centers belonging to the other classes.

\subsection{Proposed loss function}
Inspired by the center loss, in this paper we propose a new loss function for facial attributes guided sketch-photo recognition. Since in most of the available sketch datasets there is only a single pair of sketch-photo images per identity, there is no benefit in assigning a separate center to each identity as in \cite{wen2016discriminative} and \cite{qi2017contrastive}. However, in this work, we assign centers to different combinations of facial attributes. In other words, the number of centers is equal to the number of possible facial attribute combinations. To define our attribute-centered loss, it is important to briefly describe the overall structure of the recognition network.

\subsubsection{Network Structure}

\begin{figure*}
	\begin{center}
		\includegraphics[width=0.8\linewidth]{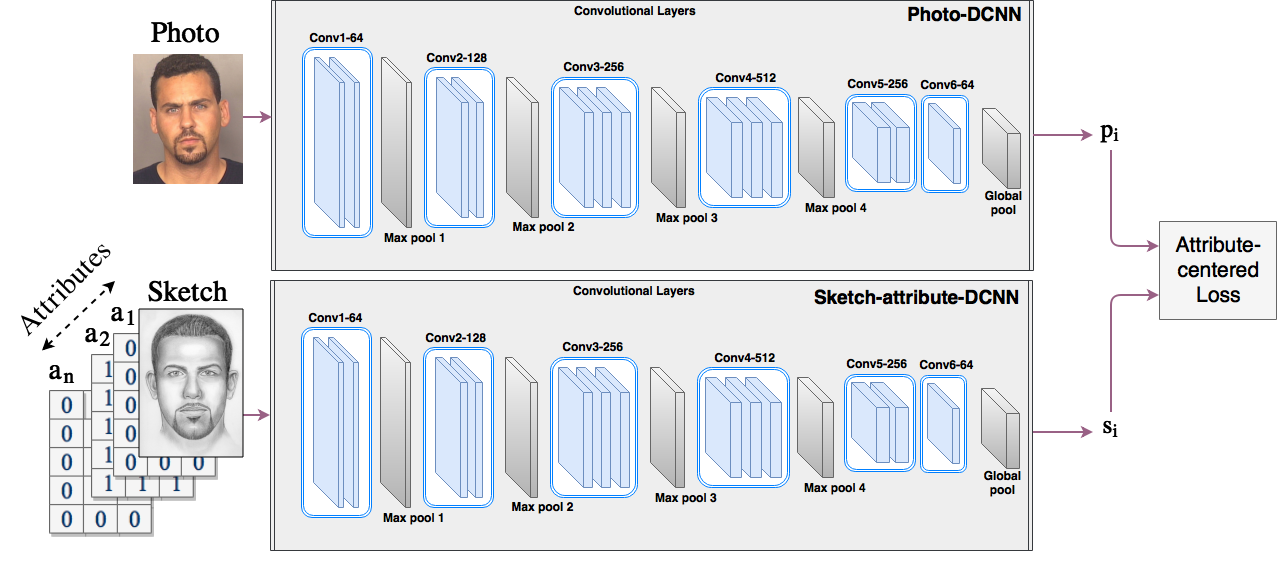}
	\end{center}
	\caption{Coupled deep neural network structure. Photo-DCNN (upper network) and sketch-attribute-DCNN (lower network) map the photos and sketch-attribute pairs into a common latent subspace.}
	\label{fig:network}
\end{figure*}

Due to the cross-modal nature of the sketch-photo recognition problem, in this work, we employed a coupled DNN model to learn a deep shared latent subspace between the two modalities, \ie, sketch and photo. Figure \ref{fig:network} shows the structure of the coupled deep neural network which is deployed to learn the common latent subspace between the two modalities. The first network, namely photo-DCNN, takes a color photo and embeds it into the shared latent subspace, $p_i$, while the second network, or sketch-attribute-DCNN, gets a sketch and its assigned class center and finds their representation, $s_i$, in the shared latent subspace. The two networks are supposed to be trained to find a shared latent subspace such that the representation of each sketch with its associated facial attributes to be as close as possible to its corresponding photo while still keeping the distance to other photos. To this end, we proposed and employed the Attribute-Centered Loss for our attribute-guided shared representation learning.

\subsubsection{Attribute-Centered Loss}

\begin{figure*}
	\begin{center}
		\includegraphics[width=0.8\linewidth]{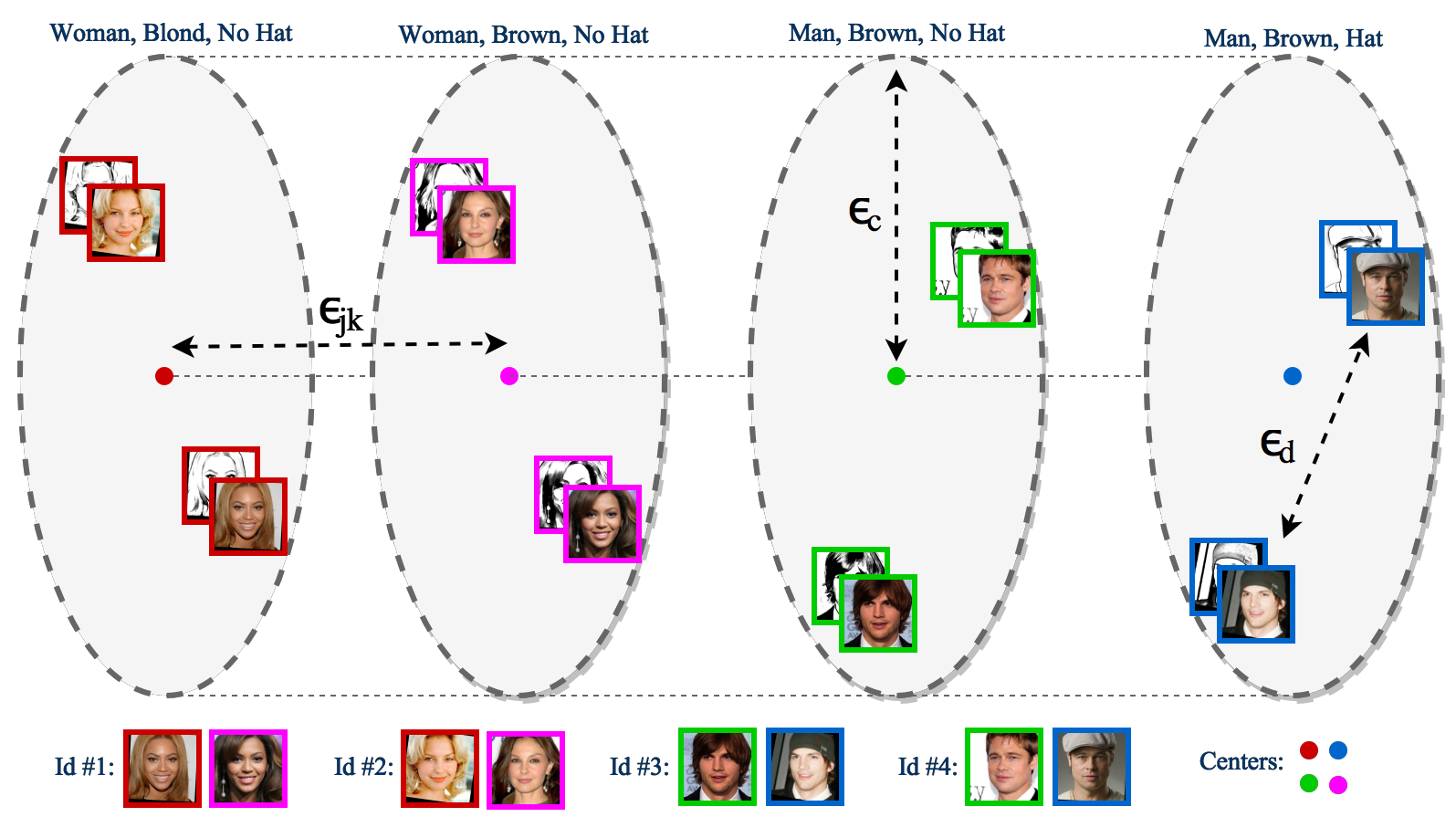}
	\end{center}
	\caption{Visualization of the shared latent space learn by the utilization of the attribute-centered loss. Centers with less contradictory attributes are closer to each other in this space.}
	\label{fig:concept}
\end{figure*}

In the problem of facial-attribute guided sketch-photo recognition, one can consider different combinations of facial attributes as distinct classes. With this intuition in mind, the first task of the network is to learn a set of discriminative features for inter-class (between different combinations of facial attributes) separability. However, the second goal of our network differs from the other two previous works \cite{wen2016discriminative, qi2017contrastive} which were looking for a compact representation of intra-class variations. On the contrary, in our work,  intra-class variations represent faces with different geometrical properties, or more specifically, different identities. Consequently, the coupled DCNN should be trained to keep the separability of the identities as well. To this end, we define the attribute-centered loss function as
\begin{align}
\label{attribute_center_loss}
L_{ac} &= L_{attr} + L_{id} + L_{cen},
\end{align}
where $L_{attr}$ is a loss to minimize the intra-class distances of photos or sketch-attribute pairs which share similar combination of facial attributes, $L_{id}$ denotes the identity loss for intra-class separability, and $L_{cen}$ forces the centers to keep distance from each other in the embedding subspace for better inter-class discrimination. The attribute loss $L_{attr}$ is formulated as  
\begin{align}
\label{loss_attr}
L_{attr} = \dfrac{1}{2}\sum_{i=1}^{m}&\max(\parallel p_i - c_{y_i} \parallel_2^2 - \epsilon_c, 0) \\ \nonumber&+ \max(\parallel s_i^g - c_{y_i} \parallel_2^2 - \epsilon_c, 0) \\ \nonumber&+ \max(\parallel s_i^{im} - c_{y_i} \parallel_2^2 - \epsilon_c, 0),
\end{align}
where $\epsilon_c$ is a margin promoting convergence, $p_i$ is the feature embedded of the input photo by the photo-DCNN with attributes combination represented by $y_i$. Also, $s_i^g$ and $s_i^{im}$ (see Figure~\ref{fig:network}) are the feature embeddings of two sketches with the same combination of attributes as $p_i$ but with the same (genuine pair) or different (impostor pair) identities, respectively. On the contrary to the center loss (\ref{center_loss}), the attribute loss does not try to push the samples all the way to the center, but keeps them around the center by a margin with a radius of $\epsilon_c$ (see figure~\ref{fig:concept}). This gives the flexibility to the network to learn a discriminative feature space inside the margin for intra-class separability. This intra-class discriminative representation is learned by the network through the identity loss $L_{id}$ which is defined as
\begin{align}
\label{loss_id}
L_{id} = \dfrac{1}{2}\sum_{i=1}^{m}&\parallel p_i - s_i^g \parallel_2^2 \\ \nonumber&+ \max(\epsilon_{d} - \parallel p_i - s_i^{im} \parallel_2^2, 0),
\end{align}
which is a contrastive loss \cite{hadsell2006dimensionality} with a margin of $\epsilon_{d}$ to push the photos and sketches of a same identity toward each other, within their center's margin $\epsilon_c$, and takes the photos and sketches of different identities apart. Obviously, the contrastive margin, $\epsilon_{d}$, should be less than twice the attribute margin $\epsilon_c$, i.e. $\epsilon_{d} < 2\times \epsilon_c$ (see Figure~\ref{fig:concept}). However, from a theoretical point of view, the minimization of identity loss, $L_{id}$, and attribute loss, $L_{attr}$, has a trivial solution if all the centers converge to a single point in the embedding space. This solution can be prevented by pushing the centers to keep a minimum distance. For this reason, we define another loss term formulated as
\begin{align}
\label{loss_cen}
L_{cen} = \dfrac{1}{2}\sum_{j=1}^{n_c} \sum_{k=1, k \neq j}^{n_c} \max(\epsilon_{jk} - \parallel c_{j} - c_{k} \parallel_2^2, 0),
\end{align}
where $n_c$ is the total number of centers, $c_j$ and $c_k$ denote the $j$th and $k$th centers, and $\epsilon_{jk}$ is the associated distance margin between $c_j$ and $c_k$. In other words, this loss term enforces a minimum distance $\epsilon_{jk}$, between each pair of centers, which is related to the number of contradictory attributes between two centers $c_j$ and $c_k$. Now, two centers which only differ in few attributes are closer to each other than those with more number of dissimilar attributes. The intuition behind the similarity-related margin is that the eyewitnesses may mis-judge one or two attributes, but it is less likely to mix up more than that. Therefore, during the test, it is very probable that the top rank suspects have a few contradictory attributes when compared with the attributes provided by the victims. Figure \ref{fig:concept} visualizes the overall concept of the attribute-centered loss. 

\subsubsection{A special case and connection to the data fusion}
For better clarification, in this section, we discuss an special case in which the network maps the attributes and geometrical information into two different subspaces. Figure \ref{fig:concept} represents the visualization of this special case. The learned common embedding space ($Z$) comprises of two orthogonal subspaces. Therefore, the basis for $Z$ can be written as the  

\begin{align}
\label{Span}
\mbox{Span}\{Z\} = \mbox{Span}\{X\}+\mbox{Span}\{Y\}, 
\end{align}
where $X\perp Y$ and $\mbox{dim}(Z) = \mbox{dim}(X)+\mbox{dim}(Y)$. In this scenario, the network learns to put the centers in the embedding subspace $X$,  and utilizes embedding subspace $Y$ to model the intra-class variations.

In other words, the learned embedding space is divided into two subspaces. The first embedding subspace represents the attribute center which provides the information regarding the subjects facial attributes. The second subspace denotes the geometrical properties of subjects or their identity information. Although this is a very unlikely scenario as some of the facial attributes are highly correlated with the geometrical property of the face, this scenario can be considered to describe the intuition behind our proposed work. 

It is important to note, the proposed attribute-centered loss guides the network to fuse the geometrical and attribute information automatically during its shared latent representation learning. In the proposed framework, the sketch-attribute-DCNN learns to fuse an input sketch and its corresponding attributes. This fusion process is an inevitable task for the network to learn the mapping from each sketch-attribute pair to its center vicinity. As shown in Figure~\ref{fig:network}, in this scheme the sketch and $n$ binary attributes, $a_{i=1, \dots, n}$, are passed to the network as a $(n+1)$-channel input. Each attribute-dedicated channel is constructed by repeating the value that is assigned to that attribute. This fusion algorithm uses the information provided by the attributes to compensate the information that cannot be extracted from the sketch (such as hair color) or it is lost while drawing the sketch.

\section{Implementation Details}
\subsection{Network Structure}
We deployed a deep coupled CNN to learn the attribute-guided shared representation between the forensic sketch and the photo modalities by employing the proposed attribute-centered loss. The overall structure of the coupled network is illustrated in Figure \ref{fig:network}. The structures of both photo and sketch DCNNs are the same and are adopted from the VGG16 \cite{simonyan2014very}. However, for the sake of parameter reduction, we replaced the last three convolutional layers of VGG16, with two convolutional layers of depth 256 and one convolutional layer of depth 64. We also replaced the last max pooling with a global average pooling, which results in a feature vector of size 64. We also added batch-normalization to all the layers of VGG16. The photo-DCNN takes an RGB photo as its input and the sketch-attribute-DCNN gets a multi-channel input. The first input channel is a gray-scale sketch and there is a specific channel for each binary attribute filled with 0 or 1 based on the presence or absence of that attribute in the person of interest. 

\subsection{Data Description}
We make use of hand-drawn sketch and digital image pairs from CUHK Face Sketch Dataset (CUFS) \cite{tang2003face} (containing 311 pairs), IIIT-D Sketch dataset \cite{bhatt2012memetic} (containing 238 viewed pairs, 140 semi-forensic pairs, and 190 forensic pairs), unviewed Memory Gap Database (MGDB) \cite{ouyang2016forgetmenot} (containing 100 pairs), as well as composite sketch and digital image pairs from PRIP Viewed Software-Generated Composite database (PRIP-VSGC) \cite{han2013matching} and extended-PRIP Database (e-PRIP) \cite{mittal2017composite} for our experiments. We also utilized the CelebFaces Attributes Dataset (CelebA) \cite{liu2015faceattributes}, which is a large-scale face attributes dataset with more than 200K celebrity images with 40 attribute annotations, to pre-train the network. To this end, we generated a synthetic sketch by applying xDOG \cite{winnemoller2012xdog} filter to every image in the celebA dataset. We selected 12 facial attributes, namely black hair, brown hair, blond hair, gray hair, bald, male, Asian, Indian, White, Black, eyeglasses, sunglasses, out of the available 40 attribute annotations in this dataset. We categorized the selected attributes into four attribute categories of hair (5 states), race (4 states), glasses (2 states), and gender (2 states). For each category, except the gender category, we also considered an extra state for any case in which the provided attribute does not exist for that category. Employing this attribute setup, we ended up with 180 centers (different combinations of the attributes). Since none of the aforementioned sketch datasets includes facial attributes, we manually labeled all of the datasets.

\subsection{Network Training}
We pre-trained our deep coupled neural network using synthetic sketch-photo pairs from the CelebA dataset. We followed the same approach as \cite{wen2016discriminative} to update the centers based on mini-batches. The network pre-training process terminated when the attribute-centered loss stopped decreasing. The final weights are employed to initialize the network in all the training scenarios. 

Since deep neural networks with a huge number of trainable parameters are prone to overfitting on a relatively small training dataset, we employed multiple augmentation techniques (see Figure~\ref{fig:augment}):
\begin{figure}
	\begin{center}
		\includegraphics[width=0.95\linewidth]{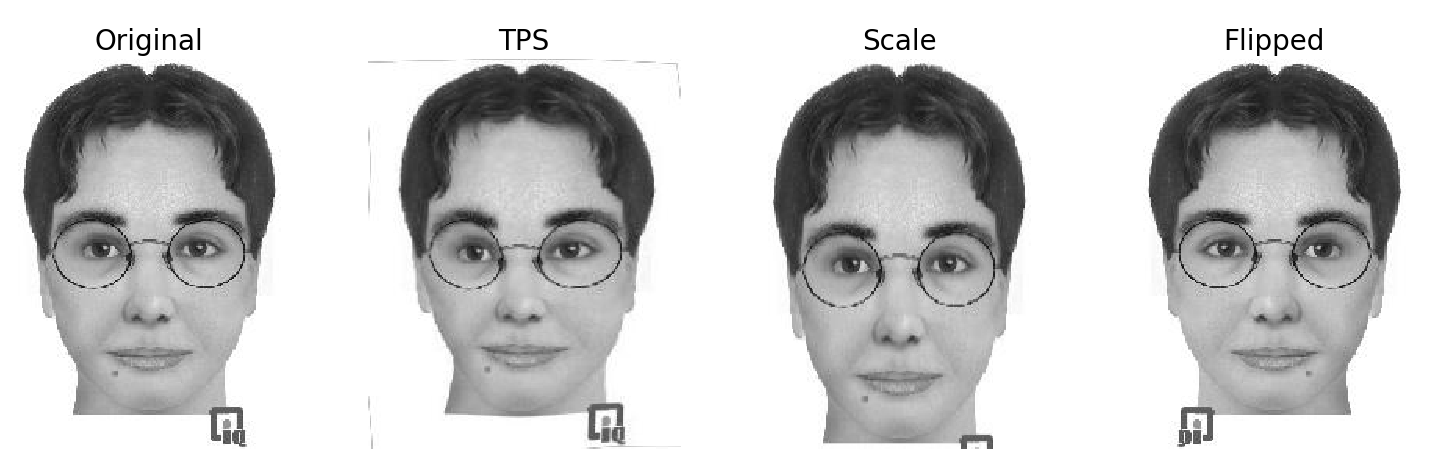}
	\end{center}
	\caption{A sample of different augmentation techniques}
	\label{fig:augment}
\end{figure}
\begin{itemize}
	\item \textbf{Deformation:} Since sketches are not geometrically matched with their photos, we employed Thin Plate Spline Transformation (TPS) \cite{bookstein1989principal} to help the network learning more robust features and prevent overfitting on small training sets, simultaneously. To this end, we  deformed images, i.e. sketches and photos, by randomly translating 25 preselected points. Each point is translated with random magnitude and direction. The same approach has been successfully applied for fingerprint distortion rectification \cite{ali18}.
	\item \textbf{Scale and crop:} Sketches and photos are upscaled to a random size, while do not keep the original width-height ratio. Then, a 250$\times$200 crop is sampled from the center of each image. This results in a ratio deformation which is a common mismatch between sketches and their ground truth photos.
	\item \textbf{Flipping:} Images are randomly flipped horizontally.
\end{itemize}

\section{Evaluation}
The proposed algorithm works with a probe image, preferred attributes and a gallery of mugshots to perform identification. In this section, we compare our algorithm with multiple attribute-guided techniques as well as those that do not utilize any extra information.
\subsection{Experiment Setup}
\begin{table*}[]
\centering
\caption{Experiment Setup}
\label{table:Ps}
\begin{tabular}{|c|c|c|c|c|c|}
\hline
Setup Name & Testing Dataset    & Training Dataset                  	& Train Size & Gallery Size & Prob Size \\ \hline
P1       & e-PRIP               & e-PRIP                            	& 48         & 75           & 75        \\ \hline
P2       & e-PRIP               & e-PRIP                            	& 48         & 1500         & 75        \\ \hline
\multirow{2}{*}{P3}  & IIIT-D Semi-forensic & \multirow{2}{*}{CUFS, IIIT-D Viewed, CUFSF, e-PRIP}    & \multirow{2}{*}{1968}       & \multirow{2}{*}{1500}         & 135       \\ \cline{2-2}\cline{6-6}
       &   MGDB Unviewed & &        &          & 100       \\ \hline
\end{tabular}
\end{table*}
We conducted three different experiments to evaluate the effectiveness of the proposed framework. For the sake of comparison, the first two experiment setups are adopted from \cite{mittal2017composite}. In the first setup, called P1, the e-PRIP dataset, with the total of 123 identities, is partitioned into training, 48 identities, and testing, 75 identities, sets. The original e-PRIP dataset, which is used in \cite{mittal2017composite}, contains 4 different composite sketch sets of the same 123 identities. However, at the time of writing of this article, there are only two of them available to the public. The accessible part of the dataset includes the composite sketches created by an Asian artist using the Identi-Kit tool, and an Indian user adopting the FACES tool. The second experiment, or P2 setup, is performed employing an extended gallery of 1500 subjects. The gallery size enlarged utilizing WVU Muti-Modal \cite{wvu}, IIIT-D Sketch, Multiple Encounter Dataset (MEDS) \cite{founds2011nist}, and CUFS datasets. This experiment is conducted to evaluate the performance of the proposed framework in confronting real-word large gallery. Finally, we assessed the robustness of the network to a new unseen dataset. This setup, P3, reveals to what extent the network is biased to the sketch styles in the training datasets. To this end, we trained the network on CUFS, IIIT-D Viewed, and e-PRIP datasets and then tested it on IIIT-D Semi-forensic pairs, and MGDB Unviewed. 

The performance is validated using ten fold random cross validation. The results of the proposed method are compared with the state-of-the-art techniques.

\subsection{Experimental results}
For the set of sketches generated by the Indian (Faces) and Asian (IdentiKit) users \cite{mittal2017composite} has the rank 10 accuracy of \%58.4 and \%53.1, respectively. They utilized an algorithm called attribute feedback to consider facial attributes on their identification process. However, SGR-DA \cite{peng2016sparse} reported a better performance of \%70 on the IdentiKit dataset without utilization of any facial attributes. In comparison, our proposed attribute-centered loss resulted in \%73.2 and \%72.6 accuracies, on Faces and IdentiKit, respectively. For the sake of evaluation, we also trained the same coupled deep neural network with the sole supervision of contrastive loss. This attribute-unaware network has \%65.3 and \%64.2 accuracies, on Faces and IdentiKit, respectively, which demonstrates the effectiveness of attributes contribution as part of our proposed algorithm.

\begin{figure}
\begin{center}
\includegraphics[width=0.95\linewidth]{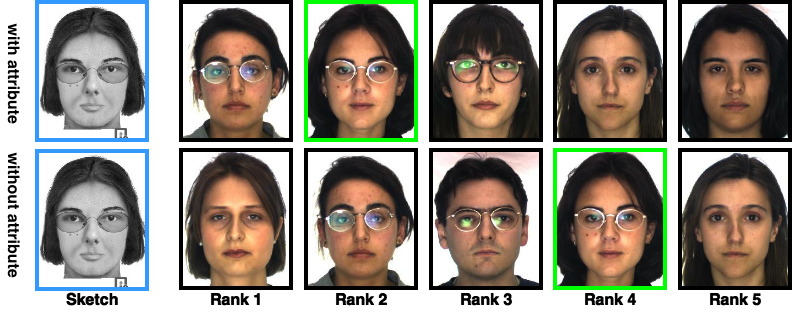}
\end{center}
   \caption{The effect of considering facial attributes in sketch-photo matching. The first line shows the results for a network trained with attribute-centered loss, and the second line depicts the result of a network trained using contrastive loss.}
\label{fig:rank_img}
\end{figure}

Figure~\ref{fig:rank_img} visualize the effect of attribute-centered loss on top five ranks on P1 experiment's test results. The first row is the results of our attribute-unaware network, while the second row shows the top ranks for the same sketch probe using our proposed network trained by the attribute-centered loss. Considering the attributes removes many of the false matches from the ranked list and the correct subject moves to a higher rank.

\begin{figure}
	\centering
	\begin{subfigure}[b]{0.45\textwidth}
		\includegraphics[width=0.94\linewidth]{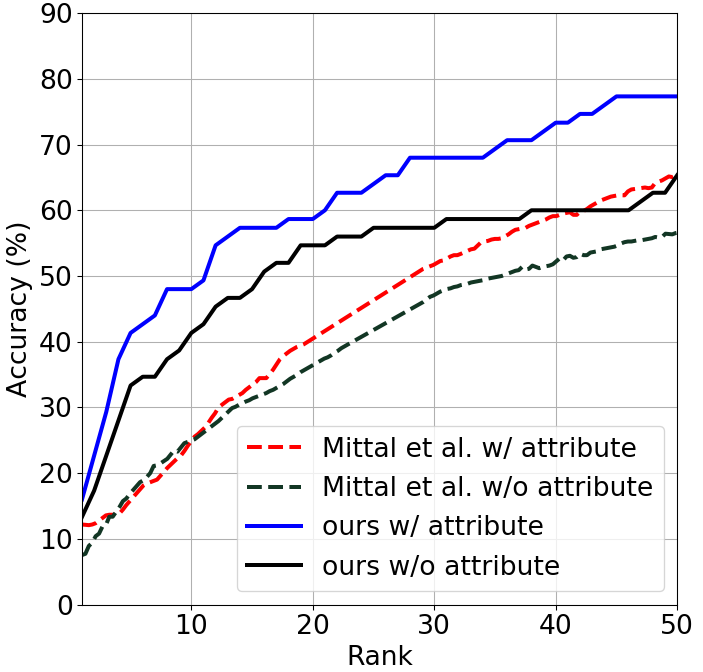}
		\caption{}
		\label{fig:ext_indian}
	\end{subfigure}
	\begin{subfigure}[b]{0.45\textwidth}
		\includegraphics[width=0.94\linewidth]{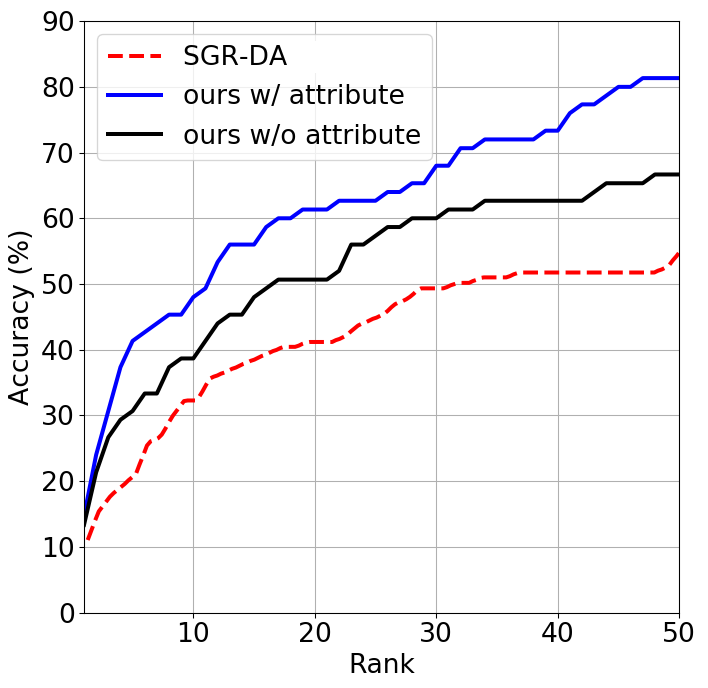}
		\caption{}
		\label{fig:ext_assian}
	\end{subfigure}
	\caption{CMC curves of the proposed and existing algorithms for the extended
		gallery experiment: (a) results on the Indian data subset compared to Mittal et al. \cite{mittal2017composite}
		and (b) results on the Identi-Kit data subset compared to SGR-DA \cite{peng2016sparse}.}\label{fig:extended}
\end{figure}

To evaluate the robustness of our algorithm in the presence of a relatively large gallery of mugshots, the same experiments are repeated but on an extended gallery of 1500 subjects. Figure~\ref{fig:ext_indian} shows the performance of our algorithm as well as the state of the art algorithm on Indian user (Faces) dataset. The proposed algorithm outperforms \cite{mittal2017composite} by almost \%11 rank 50 when exploiting facial attributes. Since the results for IdentiKit was not reported on \cite{mittal2017composite}, we compared our algorithm with SGR-DA \cite{peng2016sparse} (see Figure~\ref{fig:ext_assian}). Even tough SGR-DA outperformed our attribute-unaware network in the P1 experiment but its result was not as robust as our proposed attribute-aware deep coupled neural network.

\begin{figure}
	\begin{center}
		\includegraphics[width=0.95\linewidth]{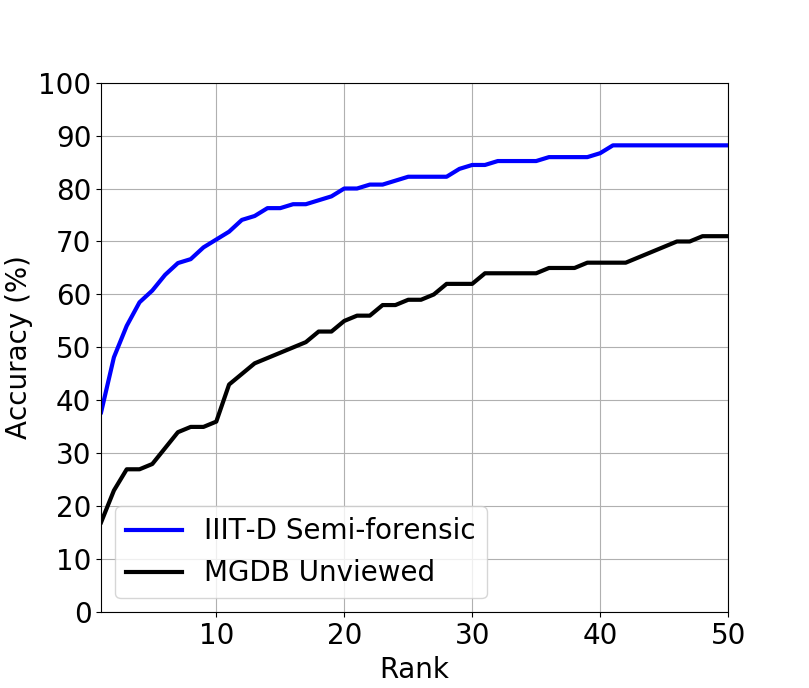}
	\end{center}
	\caption{CMC curves of the proposed algorithm for experiment P3. The results confirm the robustness of the network to different sketch styles.}
	\label{fig:p3}
\end{figure}

Finally, Figure~\ref{fig:p3} demonstrate the results of the proposed algorithm on P3 experiment. The network is trained on 1968 sketch-photo pairs and then tested on two completely unseen datasets, i.e. IIIT-D Semi-forensic and MGDB Unviewed. The gallery of this experiment was also extended to 1500 mugshots.
 
\begin{table}[]
\centering
\caption{Rank-10 identification accuracy (\%) on the e-PRIP composite sketch database.}
\label{table:P1}
\begin{tabular}{|l|c|c|}
\hline
\textbf{Algorithm}       & \textbf{Faces (In)} & \textbf{IdentiKit (As)} \\ \hline
Mittal et al. \cite{mittal2014recognizing} & 53.3 $\pm$ 1.4      & 45.3 $\pm$ 1.5              \\ \hline
Mittal et al. \cite{mittal2015composite} & 60.2 $\pm$ 2.9      & 52.0 $\pm$ 2.4              \\ \hline
Mittal et al. \cite{mittal2017composite} & 58.4 $\pm$ 1.1      & 53.1 $\pm$ 1.0              \\ \hline
SGR-DA \cite{peng2016sparse} & -     & 70              \\ \hline
Ours without attributes  & 68.6 $\pm$ 1.6           & 67.4 $\pm$ 1.9                   \\ \hline
Ours with attributes     & \textbf{73.2 $\pm$ 1.1}   &   \textbf{72.6 $\pm$ 0.9}       \\ \hline
\end{tabular}
\end{table}

\section{Conclusion}
In this work, we have proposed a novel framework to address the difficult task of cross-modal face recognition for photo and forensic sketches. By introducing an attribute-centered loss, a coupled deep neural network is trained to learn a shared embedding space between the two modalities in which both geometrical and facial attribute information cooperate on similarity score calculation. To this end, a distinct center point is constructed for every combination of the facial attributes, which are used in the sketch-attribute-DCNN, by leveraging the facial attributes of the suspect provided by the victims, and the photo-DCNN learned to map their inputs close to their corresponding attribute centers. This attribute-guided representation learning scheme helped the network to filter out the photos in the gallery that have many contradictory attributes to the attributes provided by the victim. The effectiveness of the proposed framework has been validated by extensive experiments.

{\small
\bibliographystyle{ieee}
\bibliography{egbib}
}

\end{document}